\UseRawInputEncoding

\documentclass[preprint,12pt,2025]{elsarticle}

\usepackage{amssymb}
\usepackage{rotating}   % <-- Provides the 'sideways' environment
\usepackage{booktabs}   % For professional table rules
\usepackage{multirow}   % For multirow cells in tables
\usepackage{tabularx}   % For auto-wrapping text
\usepackage[table]{xcolor} % For coloring rows and cells
\usepackage{ragged2e}   % For better text alignment in columns
\usepackage{caption}    % For captioning non-standard floats
\usepackage{subcaption}
\usepackage{makecell}
\usepackage{tabularx}
\usepackage{hyperref}

\definecolor{tableheader}{HTML}{E0F2F7} % Light blue for header
\definecolor{rowhighlight}{HTML}{F8F8F8} % Very light grey for alternating rows
\usepackage{amsmath}
\journal{Array}

\begin{document}

\begin{frontmatter}



\title{DeiTFake: Deepfake Detection Model using DeiT Multi-Stage Training} %% Article title


 %% Author name
\author[1]{Saksham Kumar}
\author[2]{Ashish Singh \corref{cor2}}
\author[3]{Srinivasarao Thota}
\author[4]{Sunil Kumar Singh\corref{cor1}}
\author[5]{Chandan Kumar}

\cortext[cor1]{Corresponding author. 
E-mail address: sunil.singh@manipal.edu (Sunil Kumar Singh)}
\cortext[cor2]{Corresponding author. 
E-mail address: ashish.singhfcs@kiit.ac.in (Ashish Singh)}
%% Author affiliations

\affiliation[1]{organization={School of Computing, Amrita Vishwa Vidyapeetham, Amaravati Campus},
            city={Amaravati},
            postcode={522503}, 
            state={Andhra Pradesh},
            country={India}}
\affiliation[2]{organization={School of Computer Engineering, Kalinga Institute of Industrial Technology (KIIT) 
Deemed to be University},
             city={Bhubaneswar},
             postcode={751024}, 
             state={Odisha},
             country={India}}
\affiliation[3]{organization={School of Engineering, Amrita Vishwa Vidyapeetham, Amaravati Campus},
            city={Amaravati},
            postcode={522503}, 
            state={Andhra Pradesh},
            country={India}}
\affiliation[4]{organization={Department of Computer Science and Engineering, 
Manipal Institute of Technology Bengaluru, 
Manipal Academy of Higher Education},
            city={Manipal},
            postcode={576104}, 
            state={Karnataka},
            country={India}}
 \affiliation[5]{organization={Amity Institute of Information Technology, Amity University Jharkhand},
            city={Ranchi},
            postcode={835303}, 
            state={Jharkhand},
            country={India}}       

%% Abstract
\begin{abstract}
Deepfakes are major threats to the integrity of digital media. We propose DeiTFake, a DeiT-based transformer and a novel two-stage progressive training strategy with increasing augmentation complexity. The approach applies an initial transfer-learning phase with standard augmentations followed by a fine-tuning phase using advanced affine and deepfake-specific augmentations. DeiT's knowledge distillation model captures subtle manipulation artifacts, increasing robustness of the detection model. Trained on the OpenForensics dataset (190,335 images), DeiTFake achieves 98.71\% accuracy after stage one and 99.22\% accuracy with an AUROC of 0.9997, after stage two, outperforming the latest OpenForensics baselines. We analyze augmentation impact and training schedules, and provide practical benchmarks for facial deepfake detection.
\end{abstract}


\begin{highlights}
\item A two-stage training approach, with progressive augmentation, is proposed for the Deepfake Detection Model
\item Used Facebook DeiT Vision Transformers for superior detection compared to existing models
\item Standard Training, followed by Fine-tuning with affine Augmentations, reached 99.22\% accuracy and 0.9997 AUROC
\end{highlights}

%% Keywords
\begin{keyword}

DeepFake Detection \sep DeiT \sep Vision Transformers \sep Transfer Learning \sep Progressive Training \sep OpenForensics
\end{keyword}


\end{frontmatter}



\section{Introduction}
\label{introduction}
%% Labels are used to cross-reference an item using \ref command.

In recent years, Generative Artificial Intelligence and Diffusion Models have made significant advances in digital media synthesis. These methodologies are democratizing synthetic media creation by making it more accessible to a wider audience. Advanced generative models have made digital media authenticity vulnerable, such as with 'Deepfakes'.

Deepfakes are increasingly used to spread misinformation and compromise individual privacy. Consequently, development of robust and updated countermeasures is essential \cite{Alanazi2024ExploringCountermeasures}. The most common alterations involve facial deepfake generation and appearance manipulations. These synthetic and altered visual media present significant challenges for digital forensics, information security, and societal trust.

Many deepfake detection models demonstrate high accuracy on controlled benchmark datasets. However, their performance often deteriorates when exposed to unseen manipulation techniques, complex processing, or entirely novel generative sources \cite{Guarnera2022TheChallenge}. Traditional detectors based on Convolutional Neural Networks (CNN), like Xception, tend to learn generator-specific artifacts, leading to fragility when faced with real-world distribution\cite{Yan2024TranscendingDetection}. Henceforth, the main challenge in digital forensics is to develop detection models that demonstrate strong generalization \cite{Abbasi2025ComprehensiveAttacks}.

\subsection{Vision Transformers and DeiT}
Recent breakthroughs in Vision Transformers(ViT) architectures, which use self-attention mechanisms to model global relationships across the image, are gaining preference for comprehensive feature extraction, as well as classification tasks \cite{Yan2023DeepfakeBench:Detection}. Given that deepfake artifacts are often subtle and have global inconsistencies (like frequency-domain anomalies, high-level semantic errors across the entire forgery), ViTs are better suited as feature extractors for forensic tasks like Deepfake Detection.

Facebook AI Research developed the Data-Efficient Image Transformer (DeiT)\cite{Touvron2020TrainingAttention}, which addressed the computational efficiency challenges in training ViT models, whilst maintaining competitive performance. DeiT uses Knowledge Distillation techniques, which accelerate the training process by data reduction, through a teacher-student strategy. The student transformer learns from the teacher network using attention layers, allowing training on standard academic datasets without the need for hundreds of millions of images.

In this work, we develop DeiTFake, a Dual-Phase Optimized DeiT Model for DeepFake Classification, which cites the generalisation gap of facial deepfake detection, using a ViT-based grounded optimization strategy.\\

Our Research Contributions are threefold:
\begin{itemize}
    \item \textbf{Two-Stage Progressive Training Framework}: A curriculum learning approach that increases augmentation and image complexity, starting with standard geometric transformations (resizing, flips, rotations), followed by advanced perturbations (ColorJitter, RandomPerspective, ElasticTransform). This strategy preserves the learned weights, improving the model’s robustness to geometric transformations.
    \item \textbf{Competitive State-of-the-Art (SOTA) Performance}: Our methodology achieves \textit{98.71\% test accuracy (F1-score:0.9871, AUROC: 0.9993)} on the OpenForensics dataset in Phase 1. Phase 2 provides additional geometric robustness through affine augmentation fine-tuning, reaching \textit{99.22\% test accuracy (F1-score:0.9922, AUROC: 0.9997)}. 
    \item \textbf{Comprehensive Methodological Analysis}: Detailed Ablation studies, architectural justification, and analysis of training strategies, relevant to deepfake detection tasks.
\end{itemize}

The remainder of this paper is structured as follows: Section 2 reviews related, recent works in deepfake detection; Section 3 outlines our methodology; Section 4 presents experimental results with a comprehensive analysis; Section 5 details our ablation study; Section 6 outlines threats to validity; Section 7 concludes with future research directions.

\section{Related Works}

\subsection{Rationale for the OpenForensics Dataset}

Earlier research in deepfake detection has been fuelled by benchmark datasets such as FaceForensics++ \cite{Rossler2019FaceForensics++:Images}, Celeb-DF v2 \cite{Li2019Celeb-DF:Forensics}, and the Deepfake Detection Challenge (DFDC) \cite{Dolhansky2020TheDataset} dataset \cite{Dolhansky2020TheDataset}. Over the past years, they have been instrumental in developing advanced detection methodologies by providing large-scale manipulated videos and images. However, they have certain limitations. FaceForensics++ and Celeb-DF are largely aligned to single-face manipulations under relatively constrained conditions, with a small number; the large DFDC dataset often contains synthetic artifacts that are less representative of real-world forgeries.

Addressing these shortcomings, more recent datasets such as DeeperForensics-1.0 \cite{Jiang2020DeeperForensics-1.0:Detection} and WildDeepfake \cite{Zi2021WildDeepfake:Detection} are used in recent studies. These datasets introduced greater diversity in manipulation techniques and environmental conditions. Still, most existing datasets are designed primarily for single‑face detection tasks. This narrow focus limits their applicability in real-world scenarios with multiple individuals appearing at a time.

The OpenForensics dataset \cite{Le2021OpenForensics:V.1.0.0} represents a significant milestone in deepfake research. It compiles a diverse collection of multi‑face forgery detection images, supporting both detection and segmentation tasks. In contrast to the earlier datasets, the OpenForensics dataset \cite{Le2021OpenForensics:V.1.0.0} captures the nuances of multi-face forgery detection and segmentation in the wild. It contains over 115,000 annotated images capturing unconstrained conditions such as varied lighting, occlusions, and complex backgrounds. Its multi-face scenarios make it more realistic in social media and surveillance contexts in real-time. This design establishes a more rigorous benchmark for evaluating deepfake detection models. By utilising global context across multiple subjects, it strengthens the model's capacity for effective deepfake detection.

\begin{table}[!htbp]
\centering
\captionof{table}{Comparison of Major Deepfake Datasets}
\label{tab:deepfake_datasets}
\begin{sideways}
\begin{tabularx}{0.95\textheight}{@{} >{\bfseries}p{3cm} 
                                        >{\RaggedRight}X 
                                        >{\RaggedRight}X 
                                        >{\RaggedRight}X 
                                        >{\RaggedRight}X 
                                        >{\RaggedRight}X @{}}
    \toprule
    \rowcolor{tableheader}
    \textbf{Dataset} & \textbf{Scale / Size} & \textbf{Media Type} & \textbf{Multi-Face Support} & \textbf{Realism / Diversity} & \textbf{Key Limitation} \\
    \midrule
    FaceForensics++ (2019) \cite {Rossler2019FaceForensics++:Images}& 1,004 videos (manipulated + pristine) & Video frames & No & Controlled conditions, multiple forgery methods & Limited diversity, mostly frontal faces \\
    \rowcolor{rowhighlight}
    Celeb-DF v2 (2020) \cite{Li2019Celeb-DF:Forensics} & 590 original + 5639 modified DF videos & Video frames & No & Celebrity-only, limited contextual diversity \\
    DFDC (2020) \cite{Dolhansky2020TheDataset} & 124,000 videos & Video frames & No & Large-scale, varied manipulations & Synthetic artifacts, less natural than in-the-wild \\
    \rowcolor{rowhighlight}
    DeeperForensics-1.0 (2020) \cite{Jiang2020DeeperForensics-1.0:Detection} & 60,000+ videos & Video frames & No & Diverse perturbations, robustness testing & Still lab-generated, not fully in-the-wild \\
    WildDeepfake (2020) \cite{Zi2021WildDeepfake:Detection} & 707 DF videos & Video frames & No & Collected from internet, more natural & Smaller scale, noisy labels \\
    \rowcolor{rowhighlight}
    OpenForensics (2021) \cite{Le2021OpenForensics:V.1.0.0} & \textbf{115,000+ images} & \textbf{Unique Original + DF Images}  & \textbf{Yes} & \textbf{In-the-wild, varied lighting, occlusion, backgrounds} & \textbf{Newer dataset, less widely adopted} \\
    \bottomrule
    \end{tabularx}
    \end{sideways}
\end{table}

\subsection{CNN Methodologies and their Limitations}

Deepfake detection research has undergone several paradigm shifts over the past decade. The field has progressed from traditional ML approaches to advanced DL architectures. Earlier methods were largely based on hand-crafted features, such as inconsistencies in facial landmarks, irregularities in eye movement patterns and photo response non-uniformity (PRNU) analysis \cite {Chai2021DeepScenarios}. Although interpretable, these hand‑crafted features offered only a limited scope. They soon became obsolete with the emerging new novel generative techniques for producing deepfake images.

CNNs have dominated the field for a large part of the research efforts \cite{Cai2023GlitchLocalization}, notably Xception, and flavours of ResNet and EfficientNet. These models typically excel at capturing fine, local artifacts in the frequency domain or specific texture distortions introduced by early-stage generative models like those found in the benchmark datasets of FaceForensics++, DFDC, and Celeb-DF. However, studies show that the model's performance degrades harshly with high False-Positive Rates (FPR) \cite{Abbasi2025ComprehensiveAttacks}, when tested on new generators and unseen degradations and manipulations, like compression or geometric distortion. Ergo, CNNs excel at feature extraction, but are computationally intensive and struggle to connect global and local pixel-level features effectively. 

\subsection{Vision Transformers for Vision Classification Tasks}

The Vision Transformers (ViT) architecture \cite{Dosovitskiy2020AnScale}, a milestone in the field of Computer Vision, applied the transformer attention mechanism \cite{Vaswani2017AttentionNeed} directly to the images. CNNs are based on convolutions to capture local spatial features, whilst ViT adopts self-attention mechanisms, treating images as sequences of patches to model long-range, interdependent global dependencies.

Although early adoption of large ViTs was limited due to resource constraints and the perceived enormous data corpus, modern approaches, like Self-Supervised Learning (SSL) pre-training techniques, have made their mark with remarkable adoption and robustness. A large number of works have shown that partial fine-tuning of pre-trained ViT models proves to be a resource-efficient alternative to foundational training from scratch, maintaining a competitive performance. Furthermore, models using SSL pre-trained backbones, such as DINO, show consistent high performance against transformations, compared to supervised backbones like CLIP. Thus, the findings back the effectiveness of Vision Transformers in classification tasks, particularly when the correlation between pixel groups is significant.

\subsection{Vision Transformers for Deepfake detection}

The application of Vision Transformers for Deepfake Detection and Segmentation is an emerging frontier in research \cite{Shelke2023AVision} \cite{Nguyen2024ExploringAnalysis} 
. Heo et al.\cite{Heo2022DeepFakeTransformer}
combined EfficientNet-B7 features with DeiT distillation strategy, showing improved detection capability. Ojha et al. \cite{Ojha2023TowardsModels} fused CLIP ViT backbones with Nearest-Neighbor Classifiers, achieving cross-dataset generalization. Cocchi et al. \cite{Cocchi2023UnveilingAnalysis} extended this work by comparing supervised CLIP with self-supervised DINO and DINOv2models. Kumar and Narang \cite{Kumar2025CombatingDetection} used a modified ViT Transformer scoring SOTA accuracies.

Related to our work on multi-face Deepfake detection using the OpenForensics dataset, only a handful of studies have been conducted so far, reflecting the relative newness of this benchmark. 

Lin et al.\cite{Lin2024ExploitingDetection} proposed \textbf{FILTER}, a framework that explicitly models correlations among multiple faces through Multi-Face Relationship Learning and Global Feature Aggregation. By capturing inter-face dependencies, FILTER achieved SOTA performance on the OpenForensics Test-Challenge split, setting a strong baseline for multi-face forgery detection. 

Building on the need for end-to-end solutions, Zhang et al. \cite{Zhang2024COMICS:Detection} introduced \textbf{COMICS}, a bi-grained contrastive learning framework that unified face extraction and forgery detection. Their approach combines coarse-grained and fine-grained contrastive objectives, subtly capturing the forgery traces across multiple faces, proving superior generalization in complex, in-the-wild scenarios. 

Gao et al.\cite{Gao2024DeepFakeContent} presented the \textbf{High-Frequency Enhancement Network (HiFE)}. HiFE emphasizes weak high-frequency cues, which are often suppressed by compression. HiFE improved robustness against low-quality and compressed content by enhancing discriminative signals. The work outperformed prior baselines on OpenForensics.

Tanfoni et al. \cite{Tanfoni2024FacialApproach} investigated \textbf{facial segmentation with transfer learning}. Pre-training on segmentation tasks improved deepfake detection by learning fine-grained structural features. As a result, the model achieves better accuracy in identifying StyleGAN2 manipulations. Their findings underscore the value of using auxiliary tasks for strengthening feature representations, with evaluations conducted on the OpenForensics sample.

\section{Methodology}

\begin{figure}
    \centering
    \includegraphics[width=1\linewidth]{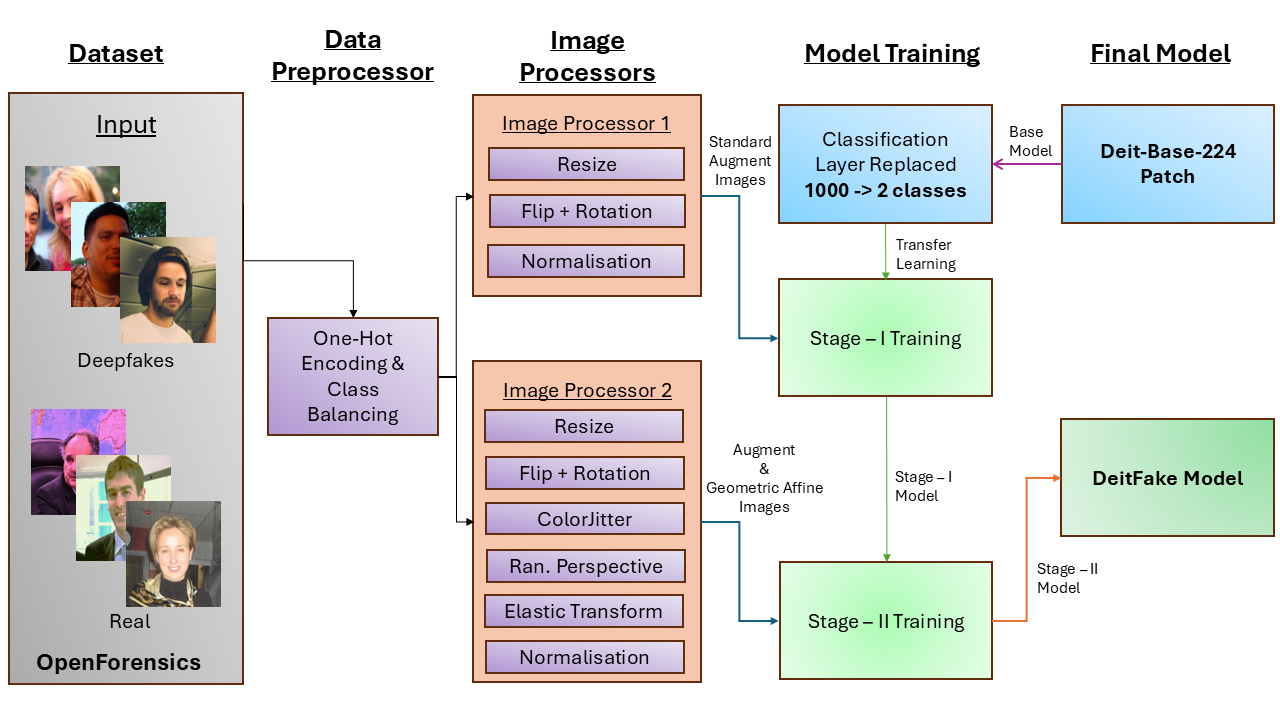}
    \caption{DeiTFake Model Pipeline}
    \label{fig:DeiTFake_pipeline}
\end{figure}

The Data-Efficient Image Transformer (DeiT) is a data-efficient variant of the Vision Transformer (ViT). It accelerates training by distillation tokens. Given the scale of our dataset, DeiT proved to be a fitting choice. Our methodology builds on these insights, using a pre-trained DeiT model and improving performance through targeted fine-tuning. The dataset is first preprocessed for class balance. The preprocessed images are passed through two separate image processors (\textit{standard} and \textit{standard+affine}), for the progressive stages of model training (Figure \ref{fig:DeiTFake_pipeline}).

\subsection{OpenForensics Dataset}

The OpenForensics dataset provides a challenging benchmark for multi-face forgery detection, comprising 190,000+ images with 334,000 face-wise annotations. Unlike conventional deepfake datasets that offer only video-level binary labels, OpenForensics includes bounding boxes, segmentation masks, forgery boundaries, and facial landmarks for each face within the images. This granular annotation supports binary classification, localization, and segmentation tasks.

For our binary deepfake detection, we utilize all OpenForensics images distributed as:
\begin{itemize}
    \item \verb|Real Faces:| 95201 (49.97\%)
    \item \verb|Fake Faces:| 94134 (50.03\%)
\end{itemize}

This perfect 1:1 balance between real and deepfake images minimizes the need for aggressive class-balancing strategies, which are common in imbalanced datasets and can cause model underfitting. 
Unlike most deepfake benchmarks (with single faces per image), OpenForensics' multi-face is still emerging as a benchmark. Our  

\subsubsection{Data Preprocessing}

\begin{figure}
    \centering
    \includegraphics[width=1\linewidth]{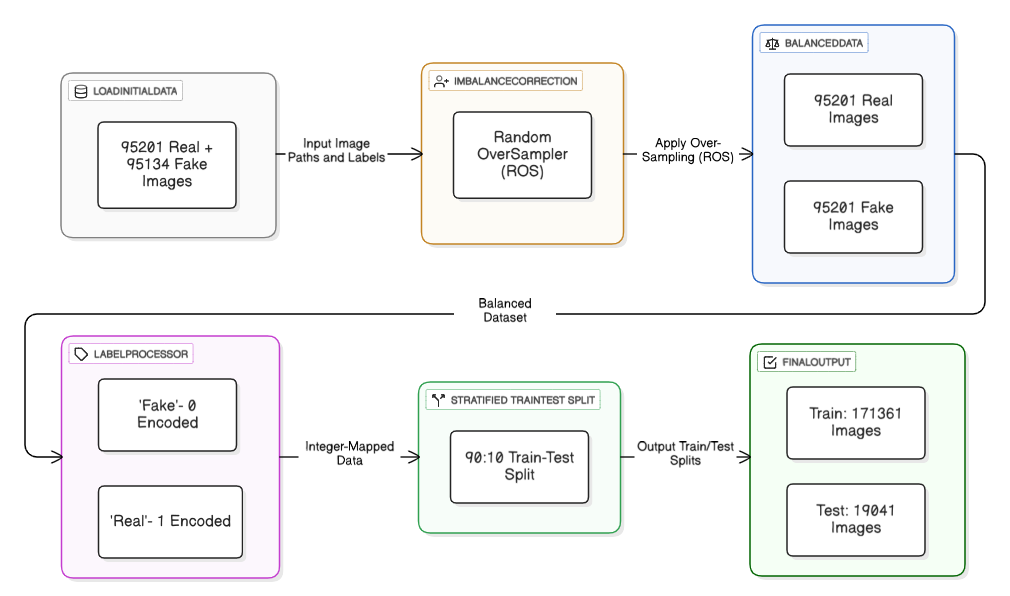}
    \caption{Common Data Preprocessor}
    \label{fig:preprocessor}
\end{figure}

We used the RandomOverSampler from the imbalanced-learn library, perfectly balancing the dataset to 95201 samples each, for a total of 190,402 samples. The dataset was stratified-split into a 9:1 ratio for train (171,361 images) and test (19,041 images) sets, using the \textit{stratify\_by\_column='label'} parameter during the \textit{train\_test\_split} operation. The stratification was necessary to maintain the real-deepfake class distribution post-dataset splitting (Figure \ref{fig:preprocessor}).

\subsection{Model Architecture}

We used the \textbf{facebook/DeiT-base-patch16-224} model as our base architecture, making use of its pre-trained ImageNet-1k weights. A summary of architectural specifications is given below:

\textbf{Image Processing:}
\begin{itemize}
    \item Input Resolution:\textit{ 224x224 pixels (RGB format)}
    \item Patch Size: \textit{16x16 pixels}
    \item Total Patches: \textit{$(\frac{224}{16})^2$ = 196 patches}
    \item Patch Embeddings Dimension: \textit{768}
\end{itemize}
\textbf{Transformer Encoder:}
\begin{itemize}
    \item Number of Layers: 12 transformer blocks
    \item Hidden Dimensions: 768
    \item Number of Attention  Heads: 12
    \item Total parameters: \textit{$ \sim $86 million}
\end{itemize}

\subsubsection {Classification Head (Modified)}

For binary deepfake classification, the pretrained ImageNet Classification Head of 1000 outputs was replaced by a binary classification head. The classification layer got an input of 768 dimensions (from the hidden layer), and output 2 classes (Fake, Real). The Softmax activation function was deemed apt for this task.

We retained the entire pre-trained DeiT backbone for Transfer Learning from ImageNet's learned visual representations. The encoder weights were preserved whilst modifying the classification layer.

\subsection{Two-Stage Progressive Training Strategy}

We implemented a curriculum-learning paradigm through two distinct stages to enhance geometric robustness and prevent catastrophic forgetting. It was done by training the model by progressively increasing the data augmentation complexity. Both the models used early stopping to avoid overfitting. 

\subsubsection{Transfer Learning with Standard Augmentations (Stage-I training)}

The model was trained for 5 epochs with a batch size of 128 images. 
We employed the \textit{AdamW} optimizer with a learning rate of $2 \times 10^{-5}$ and a weight decay of 0.01. Training was performed using FP16 mixed precision to improve computational efficiency. The model training completed in 59 minutes and 49 seconds.

Model evaluation was conducted at the end of each epoch, with the 
Area Under the Receiver Operating Characteristic (AUROC) curve 
used as the primary metric for monitoring optimization and convergence.

\textbf{Data Augmentation:}

Initially, all images were resized to $224 \times 224$ pixel resolution using bilinear interpolation, to fit the standard input dimension of the DeiT-224 architecture. The pipeline used two key geometric data augmentation techniques for training data diversity: \textit{Random Horizontal Flip} ($ p = 0.5$) and \textit{Random Rotation} (up to 15 degrees). Affine Transformations, as such, are common in image processing tasks, which allow minor shifts and distortions in real-world media, acting as implicit regularisers.

\begin{figure}
    \centering
    \includegraphics[width=1\linewidth]{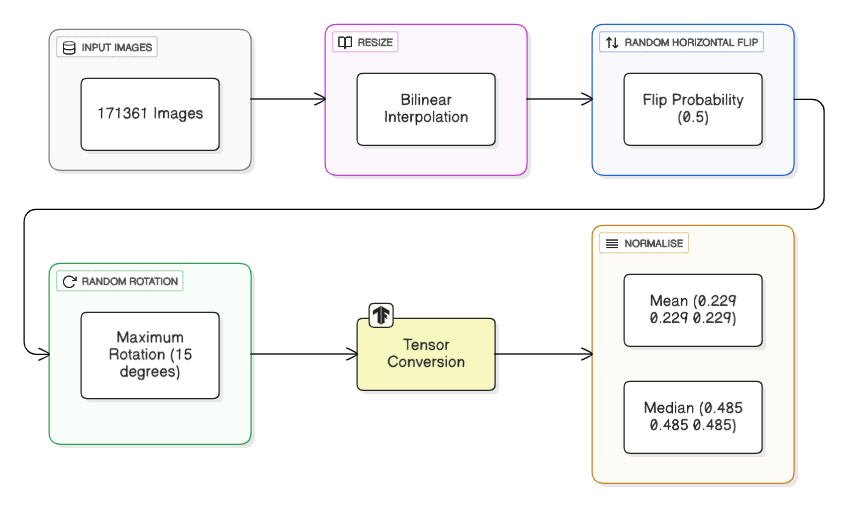}
    \caption{Stage-I Image Augmentation and Processing}
    \label{fig:stage1}
\end{figure}

Following augmentation, the images were converted to Tensors and underwent color-channel normalization. The channel-wise modification used mean ($[0.485, 0.456, 0.406]$) and standard deviation ($[0.229, 0.224, 0.225]$), derived from the ImageNet dataset. The normalisation encouraged alignment of our task dataset with the ImageNet pre-training data. The training strategy thereby maximised the benefits of the embedded features within the model, with accelerated convergence using a TPU-based environment (Figure \ref{fig:stage1}).

\textbf{Loss Function:} Cross-Entropy Loss with Softmax Outputs:
% Requires: \usepackage{amsmath}
\begin{equation}
    \mathcal{L}_{CE} = -\sum_{c=1}^{2} y_c \log(\hat{y}_c)
    \label{eq:crossentropy}
\end{equation}
where $y_c$ is the true label (one-hot encoded) and $\hat{y}_c$ is the predicted probability for class $c$.

\textbf{Optimization:} AdamW (Adam with decoupled weight decay) optimizer was used, which improves generalization with explicit L2 regularization, decoupled for gradient-based updates.

\begin{equation}
    \theta_{t+1} = \theta_t - \eta \left( \frac{m_t}{\sqrt{v_t} + \epsilon} + \lambda \theta_t \right)
    \label{eq:adamw}
\end{equation}
, where  $\eta$ is the learning rate, $m_t$ and $v_t$ are first and second moment estimates, $\lambda$ is the weight decay coefficient (0.1) and $\epsilon$ is the numerical stability constant.

\textbf{Training Dynamics}: Training progressed smoothly, with monotonic validation loss decrease and accuracy improvements across epochs.

\begin{table}[h]
    \centering
    \caption{Stage-I Training Progress}
    \label{tab:placeholder_label}
    \begin{tabular}{cccccc}
        \toprule
        Epoch & Train Loss & Val Loss & Accuracy & F1-Macro & AUROC \\
        \midrule
        1 & 0.0827 & 0.0503 & 0.9805 & 0.9805 & 0.9984 \\
        2 & 0.0454 & 0.0406 & 0.9846 & 0.9846 & 0.9990 \\
        3 & 0.0292 & 0.0384 & 0.9859 & 0.9859 & 0.9992 \\
        4 & 0.0246 & 0.0350 & 0.9877 & 0.9877 & 0.9993 \\
        5 & 0.0164 & 0.0359 & 0.9871 & 0.9871 & 0.9993 \\
        \bottomrule
    \end{tabular}
\end{table}

Training loss had consistent reductions over the epochs, from 0.0827 -> 0.0164 (80.2\% decrease), showing effective learning. Validation loss stabilizes after epoch 3, with slight oscillation (0.0384 -> 0.0350 -> 0.0359), indicating optimal regularization, without overfitting. The marginal accuracy decrease over the epochs indicates model convergence.

\subsubsection{Affine Augmentation with Color(Stage - II Training)}

Similar preprocessing techniques as Stage-I were implemented, with a few additions to enhance the model's geometric robustness. During this single-epoch retraining stage, the existing geometric augmentations were compounded with a sophisticated set of transformations to enforce invariance against real-world deepfakes. The model converged on the first epoch, in 51 minutes and 19 seconds.

\begin{figure}
    \centering
    \includegraphics[width=1\linewidth]{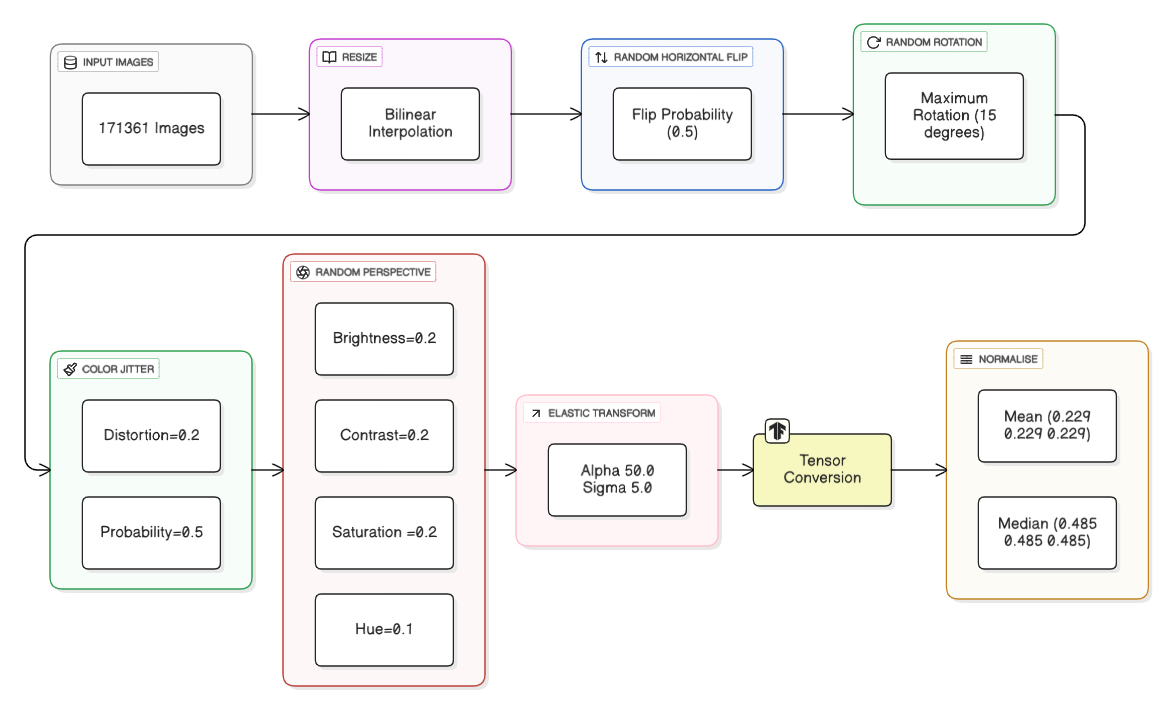}
    \caption{Stage-II Image Augmentation and Processing}
    \label{fig:stage2}
\end{figure}

Post Random Rotation, the pipeline added ColorJitter (applied to brightness, contrast, and saturation up to 20\%, and hue up to 10\%). ColorJitter added photometric robustness by countering subtle color blending and lighting inconsistencies, cropped up during image generation. Along with that, two additional geometric transformations were added: Random Perspective (distortion scale of 0.2, applied with $p=0.5$) and Elastic Transform (alpha=50.0, sigma=5.0). Random Perspective simulates non-linear warping and spatial misalignment, which happens when a 2D synthetic face is mapped on a 3D head. Elastic Transformation introduces localised, non-rigid deformations to harden the model against fine-grained warping artifacts (Figure \ref{fig:stage2}).

The comprehensive, targeted augmentation strategy helped the model to be more robust against real-world detection, retaining memory artifacts and better convergence. In the single-epoch retraining, the model got a training loss of 0.0425, validation loss og 0.0237, F1 score of 0.9919, accuracy of 0.9919, and an AUROC of 0.9997. \\

\textbf{Rationale for Affine Augmentation:}

Affine Transformations allow intricate spatial variations, and preserve essential image artifacts in the process\cite{Yu2019PDA:Networks}. Mumuni and Mumuni \cite{Mumuni2022DataApproaches} stress the relevance of data augmentation for image processing tasks, to increase the volume, quality, and diversity of training data for Machine Learning. \\ 

\textbf{Curriculum Learning Perspective}
Stage I modifies the model and teaches fundamental deepfake patterns through standard augmentations. Stage II introduces 'harder' training examples through affine transformations, refining decision boundaries for model optimization. This progressive difficulty approach prevents calamitous interference, observed on applying aggressive augmentation from initialization. 

\subsection{Training Infrastructure and Implementation}

\textbf{Hardware:} The model was trained on a \textit{Google Tensor Processing Unit (TPU) v5e-8  (8 TPU cores, 128 GB High-Bandwidth Memory (HBM)} to accelerate the computationally intensive transfer learning process with larger batch size support. The environment was crucial for stable, high-speed training of the Vision-Transformer (ViT) based DeiT model. 

\textbf{Software Framework:} We utilized the Hugging Face Transformers Library for Model Implementation, PyTorch and TorchVision for image processing and model tweaking, SciKit-Learn for dataset manipulation and evaluation.

\subsection{Evaluation Metrics}

We used multiple classification metrics, capturing performance from multiple perspectives:

\verb|Accuracy|: Overall Classification correctness across both classes (Real and Deepfake:

\begin{equation}
    \text{Accuracy} = \frac{TP + TN}{TP + TN + FP + FN}
\end{equation} \\

\verb|Macro-Averaged F1 Score|: Harmonic mean of precision and recall, averaged across classes without weighting by class size. Used for capturing small-class imbalances:

\begin{equation}
    F_{1\ \mathrm{macro}} = \frac{1}{2} \sum_{c=1}^2 \frac{2 \cdot \mathrm{Precision}_c \cdot \mathrm{Recall}_c}{\mathrm{Precision}_c + \mathrm{Recall}_c}
\end{equation}
This metric provides a balanced assessment in cases of slight class imbalance. \\

\verb|AUROC (Area Under Receiver Operating Characteristic):|AUROC measures the model's ability to distinguish between classes across all classification thresholds. AUROC is considered vital for deployment scenarios requiring threshold tuning, especially in security contexts:

\begin{equation}
    \text{AUROC} = \int_{0}^{1} \text{TPR}(t)\, d\text{FPR}(t)
\end{equation}

where TPR (True Positive Rate) = $TP/(TP+FN)$ and FPR (False Positive Rate) = $FP/(FP+TN)$. AUROC values approaching 1.0 indicate excellent discriminative ability, while 0.5 represents random guessing.
\section{Experiment Results}

\subsection{Overall Performance Metrics \& Inferencing}
\begin{figure}
    \centering
    \includegraphics[width=1\linewidth]{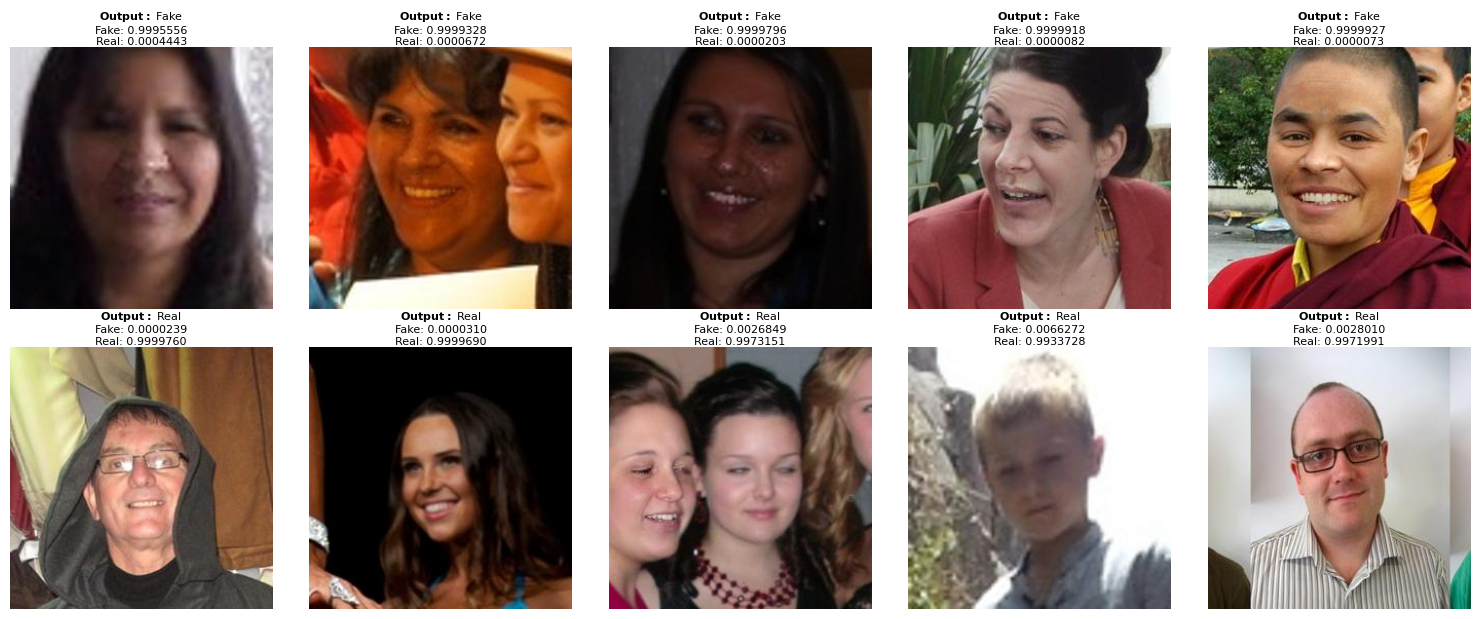}
    \caption{Inference Results on Test Images}
    \label{fig:inference}
\end{figure}

The Dual-Phase Optimised DeiTFake model achieved a new high mark on the OpenForensics dataset. Specifically, the model reached an overall accuracy of 0.9922, a Macro F1-score of 0.9922, and an exceptional AUC-ROC score of 0.9997 on the testing partition, with the dissuaded results showcased in table \ref{tab:test_results}. The raw images were classified flawlessly with an accuracy of 99\%+, as shown in Figure \ref{fig:inference}.

\begin{table}[h!]
    \centering
    
    \begin{tabular}{lcc}
    \hline
    \textbf{Metric} & \textbf{Stage-I} & \textbf{Stage-II} \\
    \hline
        Test Loss      & 0.03588  & 0.02191 \\
        Test Accuracy  & 0.98713  & 0.99223 \\
        Test F1 Macro  & 0.98713  & 0.99223 \\
        Test AUROC     & 0.99933  & 0.99973 \\
        % Support        & 9,520    & 9,521   
    \hline
    \end{tabular}
    \caption{Comparison of Metrics for Stage-I and Stage-II}
    \label{tab:test_results}
\end{table}

\begin{figure}[htbp]
    \centering
    
    % --- Top Row ---
    % Subfigure (a): Confusion Matrix (Stage-I)
    \subfloat[Confusion Matrix (Stage-I)]{\includegraphics[width=0.45\linewidth]{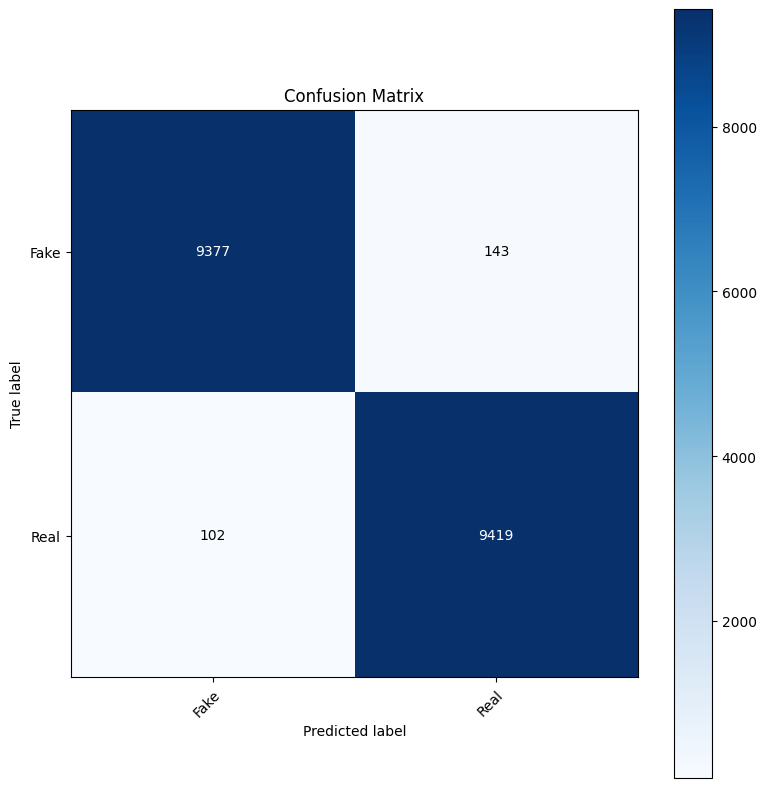}\label{fig:CF1}}
    \hfill % Adds horizontal space between the two subfigures
    % Subfigure (b): Confusion Matrix (Stage-II)
    \subfloat[Confusion Matrix (Stage-II)]{\includegraphics[width=0.45\linewidth]{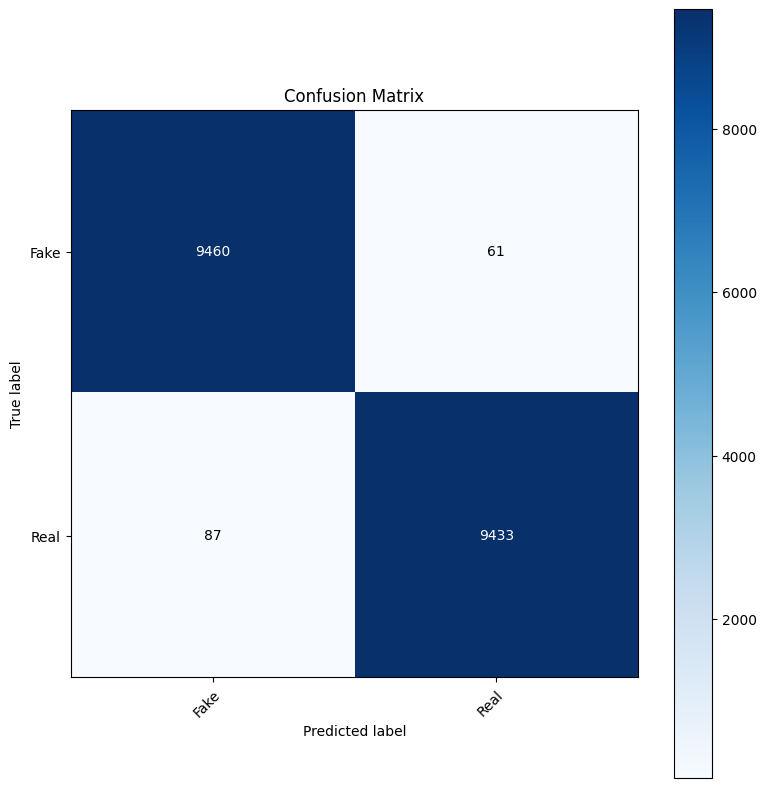}\label{fig:CF2}}
    \\ % Starts a new row
    
    % --- Bottom Row ---
    % Subfigure (c): ROC Curve (Stage-I)
    \subfloat[ROC Curve (Stage-I)]{\includegraphics[width=0.45\linewidth]{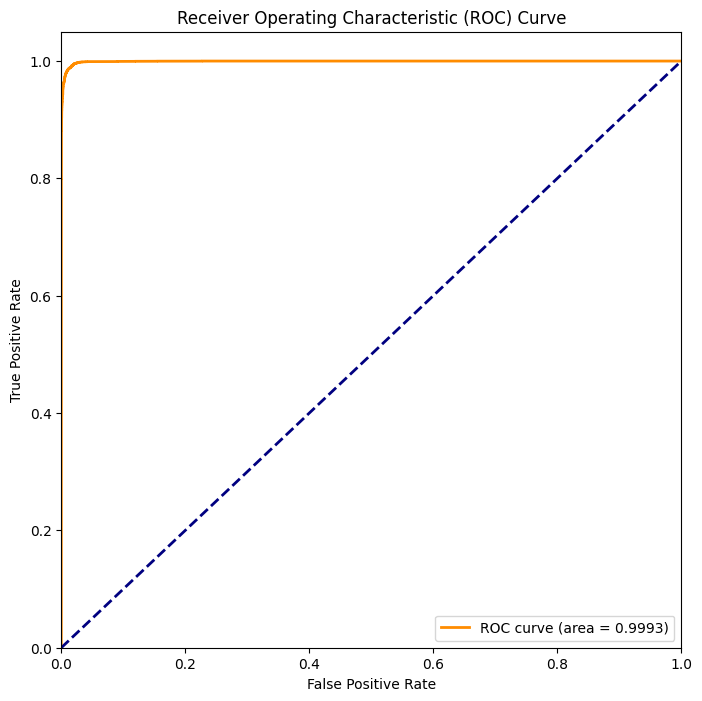}\label{fig:ROC1}}
    \hfill
    % Subfigure (d): ROC Curve (Stage-II)
    \subfloat[ROC Curve (Stage-II)]{\includegraphics[width=0.45\linewidth]{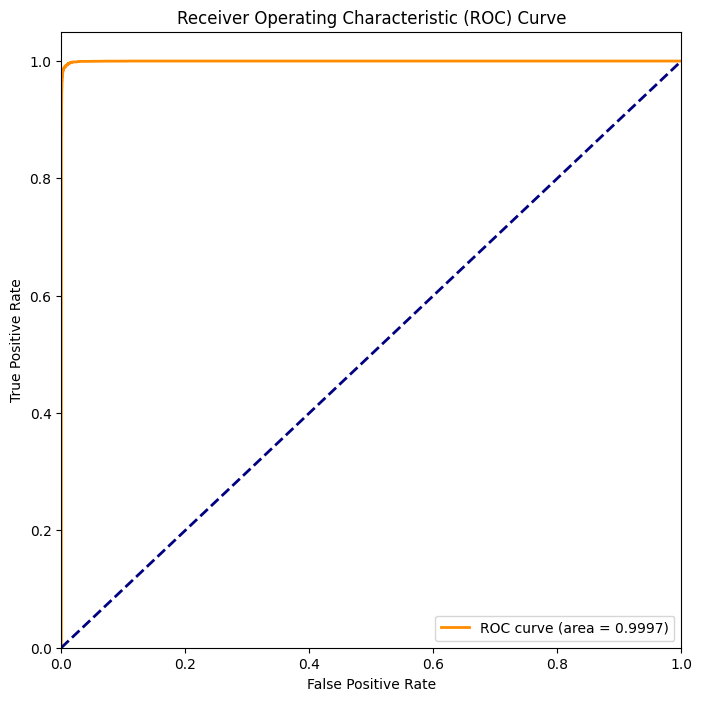}\label{fig:ROC2}}

    % Main Caption for the entire figure
    \caption{Combined visualization of model performance metrics. Figures (a) and (b) show the Confusion Matrices for Stage-I and Stage-II, while (c) and (d) display the corresponding ROC Curves.}
    \label{fig:combined_metrics}
\end{figure}

\subsection{Key Observations}
\begin{enumerate}
    \item \textbf{Symmetric Performance amongst classes:} A near-identical performance across both classes \textit{(Fake F1: 0.9871 vs. Real F1: 0.9872)}, indicating no class bias.
    \item \textbf{High-Precision Recall Balance:} Both fake and real classes maintain precision and recall above 98.5\%. Therefore, the model avoids common pitfalls of precision-recall trade-offs.
    \item \textbf{Excellent AUROC: } The 0.9993 AUROC indicates the model produces well-separated distributions for fake and real classes. It is necessary for deployment-critical models, where the threshold can be flexible (Figure \ref{fig:combined_metrics}).
    \item \textbf{Low False Negative Rate:} 143 fake faces are misclassified as real (FNR = 143/9520 = 1.50 \%) remains acceptably low for security applications.
    
\end{enumerate}

\subsection{Comparison with Previous works}

\begin{figure}
    \centering
    \includegraphics[width=1\linewidth]{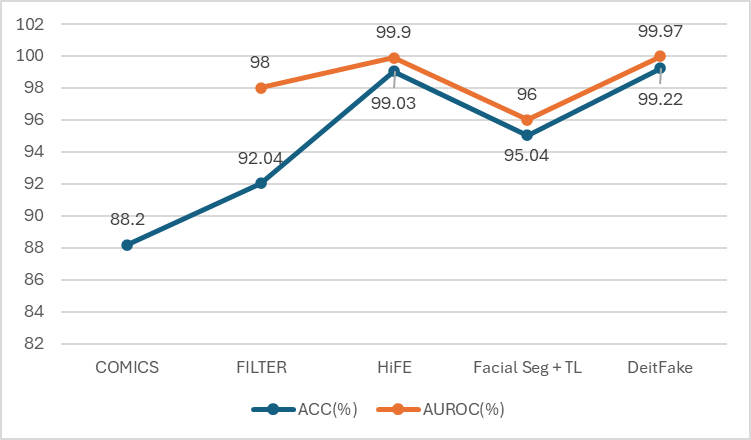}
    \caption{Comparison of Works based on Accuracy and AUROC}
    \label{fig:comparative_plot}
\end{figure}

Table \ref{tab:comparative_survey} summarizes the recent novel works in the field of Deepfake Detection using the OpenForensics dataset. The comparative results illustrate the methodological diversity, ranging from relationship modelling and contrastive learning to frequency enhancement and segmentation transfer, despite all being evaluated on the same dataset. A comparative line plot (Figure \ref{fig:comparative_plot}) provides a visual representation of the compared models. 

\begin{table}[!htbp]
\centering
\caption{Comparison of Deepfake detection methods on the OpenForensics dataset's test set.}
\label{tab:comparative_survey}
\begin{tabular}{|p{3.5cm}|p{4.5cm}|c|c|}
\hline
\textbf{Work} & \textbf{Methodology} & \textbf{Accuracy (\%)} & \textbf{AUROC} \\
\hline
Zhang et al. (2024) -- COMICS \cite{Zhang2024COMICS:Detection} & Bi-grained contrastive learning with integrated face extraction & 88.20 & - \\
\hline
Lin et al. (2024) -- FILTER \cite{Lin2024ExploitingDetection} & Multi-face relationship learning + global feature aggregation & 92.04 & 0.9800 \\
\hline

Tanfoni et al. (2024) \cite{Tanfoni2024FacialApproach} & Facial Segmentation + transfer learning (DeepLabV3+) & 95.04 & 0.9600 \\
\hline
Gao et al. (2024) -- HiFE \cite{Gao2024DeepFakeContent} & High-frequency enhancement network for compressed content & 99.03 & 0.9990 \\
\hline
\textbf{\textit{DeiTFake : DPO-DeiT-based Model}} & \textbf{\textit{Dual-Stage Trained DeiT Model with Complex Augmentations}} & \textbf{\textit{99.22}} & \textbf{\textit{0.9997}} \\
\hline
\end{tabular}
\end{table}

\section{Ablation Study: Validation of Dual-Phase Optimization and Affine Transformations}

The ablation tests were designed to confirm the efficacy of the three methodological choices - Resource-Efficient Partial Fine-tuning, the structural benefit of the generalization (compression) phase, and the functional improvement derived from imposing geometric invariance (via affine transformations).

\subsection{ Component I - DeiT Architectural Adaptation}

\begin{figure}
    \centering
    \includegraphics[width=1\linewidth]{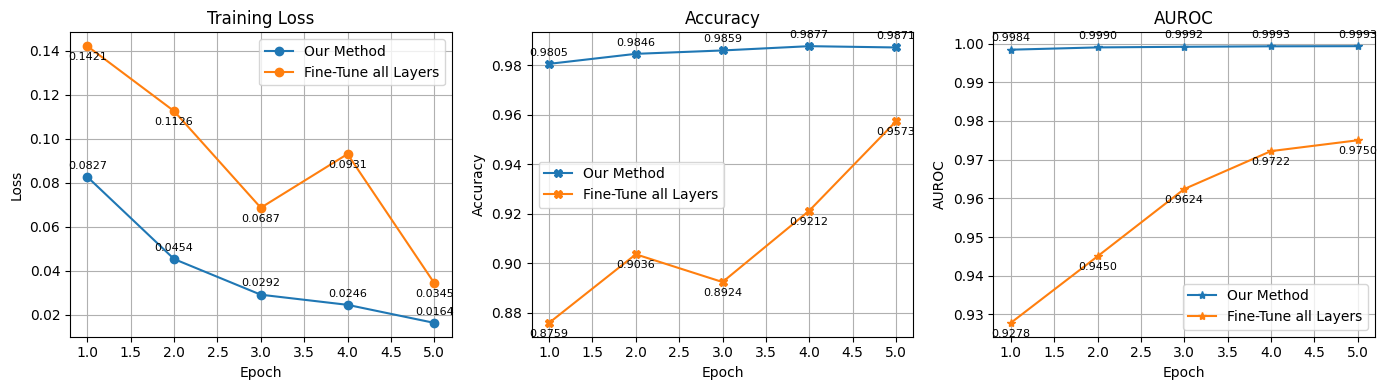}
    \caption{Comparison of DeiTFake Architectural Training  (Layer Freezing)}
    \label{fig:ablation1}
\end{figure}

For validating the efficient fine-tuning approach, a comparison was made between the chosen strategy (initial layers frozen + fine-tuning later blocks) and a baseline of fine-tuning all the layers. The results confirmed that partial fine-tuning of the deeper layers was the optimal strategy, as the latter faced multiple lags and instability of optimization during the training phase. It might be a result of forgetting the original ImageNet extraction artifacts, which might have decayed during fine-tuning of all layers, at once. The comparative results are shown in Figure \ref{fig:ablation1}.

\subsection{Component - II - Efficacy of Dual-Phase Training Structure}

The study isolated the effect of the sequential, compressed training period, independent of the affine transforms.
\begin{itemize}
    \item \textbf{Test Case T1 (Baseline Fit): } Model trained for 5 epochs (Phase - I) with standard augmentation only.
    \item \textbf{Test Case T2 (Structural Compression): } Model trained for 5 epochs, then retrained for 1 additional epoch  (Phase-II) using only standard transformations.
\end{itemize}

\textbf{T2} achieved a slightly generalised gain (High AUC-ROC) compared to \textbf{T1}. Therefore, the transition to a multi-stage approach structurally improves the model's ability to generalize, finding a more stable minimum in the loss landscape.

\subsection{Component III - Targeted Affine Transformations}

The final model, \textbf{T3} (Model trained for 5 epochs (Phase I), then retrained for 1 epoch (Phase II), with Standard + Affine transforms) quantified the specific performance bumps due to geometric robustness.

The comparison between T1 (structural compression only), T2 (structural compression in dual phase) and T3 (Dual Phase with Geometric Transforms) , showed a measurable increase in both accuracy and AUC-ROC. The results thereby validate the effectiveness and contribution of complex geometric transformations, which improves upon generalization in modern deepfake benchmarks. In conclusion, the findings validate the hypothesis that affine transforms effectively mitigate warping artifacts in deepfakes (Table \ref{tab:Ablationstudy}).

\begin{table}[ht]
    \centering
    \small
    \renewcommand{\arraystretch}{1.2}
    \begin{tabularx}{\textwidth}{c p{2.2cm} c c c c c}
        \hline
        \makecell{Test \\ Case} & \makecell{Architectural \\ Modification} &
        \makecell{Total \\ Epochs} & \makecell{Affine \\ Transform \\ Used?} &
        \makecell{Final \\ Accuracy (\%)} & \makecell{Final \\ Macro F1} &
        \makecell{Final \\ AUC-ROC} \\
        \hline
        T1 & Baseline Fit (S1 with Std. Aug.) & 5 & No & 95.29 & 0.9479 & 0.9222 \\
        T2 & Stage-I\&II (Standard Augmentations) & 5+1 & No & 98.71 & 0.9781 & 0.9894 \\
        \textbf{T3} & \textbf{DeiTFake (Complex Augmentations, Dual Layers)} & 5+1 & Yes & \textbf{99.22} & \textbf{0.9922} & \textbf{0.9997} \\
        \hline
    \end{tabularx}
    \caption{Ablation study results with architectural modifications and training settings.}
    \label{tab:Ablationstudy}
\end{table}

\section{Threats to Validity}
This study acknowledges several potential threats to the validity of our empirical evaluation.
External validity concerns the generalizability of the results to other deepfake datasets and generation methods beyond OpenForensics - diffusion models (e.g., Stable Diffusion), NeRF-based re-enactment, and other advanced GAN variants. These may produce distinct artifacts not encountered during training. 

Internal validity is supported by rigorous training protocols, an ablation study (Section 6), and reproducibility measures. Nonetheless implementation choices, such as optimizer parameters and augmentation schedules, can influence outcomes.

Construct validity is addressed by adopting standard, well-accepted metrics for deepfake studies - Accuracy, Macro F1, and AUROC. Still, these metrics may not capture model behaviours under adversarial attacks or domain shifts. The reliance on single-frame image inputs also limits the temporal consistency. It probably affects detectability in video sequences, useful in real-world scenarios.

Consequently, the inherent black-box nature of usual Deep Learning models, including DeiTFake, constrains interpretability. Attention mechanisms suggest focus on manipulated pixel-zones, yet formal explainability analyses have yet to be conducted. The demanding architecture of DeiT might also limit accessibility in edge deployments.

\section{Conclusion and Future Work}

The Dual-Phase Optimized DeiT Deepfake Classifier (\textbf{DeiTFake}) is a highly effective and resource-efficient classifier for deepfake detection. Using global feature extraction of DeiT with dual-phase optimization, the model gets a competitive performance. The performance improvement observed during Stage-II training was validated by the ablation study, confirming the efficacy of complex geometric transformations. In particular, the affine transforms strengthened the detector against common face-warping artifacts, frequently produced by deepfake generators.

DeiTFake's innovative methodology outlines an optimal blueprint for future classification systems, not just deepfake detectors. By strategically separating the learning process into 'acquisition' and 'generalization' phases, researchers can fine-tune ViT backbones more effectively. In the context of deepfakes, models can prioritize stable forensic artifacts. They do not rely only on generator‑specific noise. This makes detectors less brittle. It also makes them more reliable in real‑world deployments. Novel generators and attack strategies are often encountered in such settings.

The principal limitation of this study is its focus on \textit{facial deepfakes}. The model’s applicability to broader modalities remains unexplored, including full‑body manipulation, audio synthesis, and lip‑syncing.

Another limitation concerns resilience against advanced adversarial perturbations. These attacks are specifically designed to evade detection, and further testing is required for deployment feasibility. Addressing this gap is crucial, ensuring reliable performance in practical settings.\\

\textbf{Future Research Directions:}
\begin{enumerate}
    \item \textbf{Architectural Optimization: } Focus on Vision Transformer Variants like Swin Transformer (hierarchical attention), Pyramid Vision Transformer, DeiT-III for better efficiency. Development of Hybrid architectures combining ViT (Global Feature Extraction) and Neural Nets (Local Feature Extraction). Techniques like knowledge distillation, pruning, or quantization for deployment optimization while maintaining accuracy.
    \item \textbf{Multi-Modal Detection: } Integrate audio-visual fusion techniques for deepfake detection (voice cloning, TTS) with visual analysis. Implement associated textual context (captions, metadata) for holistic content authenticity assessment.
    \item \textbf{Explainability and Trust: } Employ attention visualisation methods such as GradCAM, attention rollout, or LIME (Local Interpretable Model-agnostic Explanations) to generate human-interpretable explanations, increasing trust in the systems.
    \item \textbf{Benchmark Expansion: } Conduct Cross-Dataset Evaluation by testing on FaceForensics++, DFDC, Celeb-DF, and DeepForgeries to assess generalization. Evaluate performance against cross-generation outputs, specifically diffusion models, NeRF outputs, and other post-2021 generation techniques not present in training data. Conduct a real-world deployment study with field evaluation on user-generated content platforms to assess practical performance. \\
\end{enumerate}

\section*{Declarations}
\subsection*{Model Availability} In the spirit of open science, we have shared the DeiTFake model on the Hugging Face Model Hub, making it openly accessible to the research community. 
\textbf{Model Link:} \href{https://doi.org/10.57967/hf/6767}{\textit{DeiTFake | HuggingFace}}.

Open availability of the model aligns with our principles for democratising access to research. We also invite improvements and adaptation, strengthening the collective defense against evolving digital manipulations. \\
\subsection*{Funding}
No specific grants or funding support.
\subsection*{Data availability statement}
Cited in the manuscript. Publicly available.
\subsection*{Conflict of interest}
No conflicts of interest.
\subsection*{Human participants and/or animals} This article does not contain human or animal studies.
\subsection*{Informed consent}
Not applicable.

%%%%%%%%%%% AI Declaration %%%%%%%%%%%%%%
\subsection*{\textbf{Declaration of generative AI and AI-assisted technologies in the manuscript preparation process:}} During the preparation of this work, the author(s) used Perplexity and Gemini to assist in structuring and refining article sections, including grammatical checks and improvements to context flow. After using these tools, the author(s) reviewed and edited the content as necessary and assume full responsibility for the final version of the published article.

% \appendix
% \section{Example Appendix Section}


% %% For citations use: 
% %%       \citet{<label>} ==> Lamport (1994)
% %%       \citep{<label>} ==> (Lamport, 1994)
% %%
% Example citation, See \citet{lamport94}.

\newpage
\bibliographystyle{apalike} 
\bibliography{references}

@article{Shelke2023AVision,
    title = {{A Review Paper on Computer Vision}},
    year = {2023},
    journal = {International Journal of Advanced Research in Science, Communication and Technology (IJARSCT)},
    author = {Shelke, Shreya M and Pathak, Indrayani S and Sangai, Aniket P and Lunge, Dipali V and Shahale, Kalyani A and Vyawahare, Harsha R},
    number = {2},
    volume = {3},
    url = {www.ijarsct.co.in},
    doi = {10.48175/IJARSCT-8901},
    issn = {2581-9429}
}

@article{Dosovitskiy2020AnScale,
    title = {{An Image is Worth 16x16 Words: Transformers for Image Recognition at Scale}},
    year = {2020},
    journal = {ICLR 2021 - 9th International Conference on Learning Representations},
    author = {Dosovitskiy, Alexey and Beyer, Lucas and Kolesnikov, Alexander and Weissenborn, Dirk and Zhai, Xiaohua and Unterthiner, Thomas and Dehghani, Mostafa and Minderer, Matthias and Heigold, Georg and Gelly, Sylvain and Uszkoreit, Jakob and Houlsby, Neil},
    month = {10},
    publisher = {International Conference on Learning Representations, ICLR},
    url = {https://arxiv.org/pdf/2010.11929},
    arxivId = {2010.11929}
}

@article{Vaswani2017AttentionNeed,
    title = {{Attention Is All You Need}},
    year = {2017},
    journal = {ArXiV},
    author = {Vaswani, Ashish and Brain, Google and Shazeer, Noam and Parmar, Niki and Uszkoreit, Jakob and Jones, Llion and Gomez, Aidan N and Kaiser, Łukasz and Polosukhin, Illia},
    month = {6},
    pages = {1},
    url = {https://arxiv.org/pdf/1706.03762},
    isbn = {1706.03762v7},
    arxivId = {1706.03762}
}

@article{Li2019Celeb-DF:Forensics,
    title = {{Celeb-DF: A Large-scale Challenging Dataset for DeepFake Forensics}},
    year = {2019},
    journal = {Proceedings of the IEEE Computer Society Conference on Computer Vision and Pattern Recognition},
    author = {Li, Yuezun and Yang, Xin and Sun, Pu and Qi, Honggang and Lyu, Siwei},
    month = {9},
    pages = {3204--3213},
    publisher = {IEEE Computer Society},
    url = {https://arxiv.org/pdf/1909.12962},
    doi = {10.1109/CVPR42600.2020.00327},
    issn = {10636919},
    arxivId = {1909.12962}
}

@article{Kumar2025CombatingDetection,
    title = {{Combating Digitally Altered Images: Deepfake Detection}},
    year = {2025},
    journal = {ArXiV},
    author = {Kumar, Saksham and Narang, Rhythm},
    month = {8},
    url = {http://arxiv.org/abs/2508.16975},
    arxivId = {2508.16975},
    language = {English}
}

@article{Zhang2024COMICS:Detection,
    title = {{COMICS: End-to-End Bi-Grained Contrastive Learning for Multi-Face Forgery Detection}},
    year = {2024},
    journal = {IEEE Transactions on Circuits and Systems for Video Technology},
    author = {Zhang, Cong and Qi, Honggang and Wang, Shuhui and Li, Yuezun and Lyu, Siwei},
    number = {10},
    pages = {10223--10236},
    volume = {34},
    publisher = {Institute of Electrical and Electronics Engineers Inc.},
    doi = {10.1109/TCSVT.2024.3405563},
    issn = {15582205}
}

@article{Abbasi2025ComprehensiveAttacks,
    title = {{Comprehensive Evaluation of Deepfake Detection Models: Accuracy, Generalization, and Resilience to Adversarial Attacks}},
    year = {2025},
    journal = {Applied Sciences 2025, Vol. 15,},
    author = {Abbasi, Maryam and V{\'{a}}z, Paulo and Silva, José and Martins, Pedro and Abbasi, Maryam and V{\'{a}}z, Paulo and Silva, José and Martins, Pedro},
    number = {3},
    month = {1},
    volume = {15},
    publisher = {Multidisciplinary Digital Publishing Institute},
    url = {https://www.mdpi.com/2076-3417/15/3/1225},
    doi = {10.3390/APP15031225},
    issn = {2076-3417}
}

@article{Mumuni2022DataApproaches,
    title = {{Data augmentation: A comprehensive survey of modern approaches}},
    year = {2022},
    journal = {Array},
    author = {Mumuni, Alhassan and Mumuni, Fuseini},
    month = {12},
    pages = {100258},
    volume = {16},
    publisher = {Elsevier},
    url = {https://www.sciencedirect.com/science/article/pii/S2590005622000911},
    doi = {10.1016/J.ARRAY.2022.100258},
    issn = {2590-0056}
}

@article{Chai2021DeepScenarios,
    title = {{Deep learning in computer vision: A critical review of emerging techniques and application scenarios}},
    year = {2021},
    journal = {Machine Learning with Applications},
    author = {Chai, Junyi and Zeng, Hao and Li, Anming and Ngai, Eric W.T.},
    month = {12},
    pages = {100134},
    volume = {6},
    publisher = {Elsevier},
    url = {https://www.sciencedirect.com/science/article/pii/S2666827021000670},
    doi = {10.1016/J.MLWA.2021.100134},
    issn = {2666-8270}
}

@article{Jiang2020DeeperForensics-1.0:Detection,
    title = {{DeeperForensics-1.0: A Large-Scale Dataset for Real-World Face Forgery Detection}},
    year = {2020},
    journal = {Proceedings of the IEEE Computer Society Conference on Computer Vision and Pattern Recognition},
    author = {Jiang, Liming and Li, Ren and Wu, Wayne and Qian, Chen and Loy, Chen Change},
    month = {1},
    pages = {2886--2895},
    publisher = {IEEE Computer Society},
    url = {https://arxiv.org/pdf/2001.03024},
    doi = {10.1109/CVPR42600.2020.00296},
    issn = {10636919},
    arxivId = {2001.03024}
}

@article{Heo2022DeepFakeTransformer,
    title = {{DeepFake detection algorithm based on improved vision transformer}},
    year = {2022},
    journal = {Applied Intelligence},
    author = {Heo, Young Jin and Yeo, Woon Ha and Kim, Byung Gyu},
    number = {7},
    month = {4},
    pages = {7512--7527},
    volume = {53},
    publisher = {Springer USNew York},
    url = {https://dl.acm.org/doi/10.1007/s10489-022-03867-9},
    doi = {10.1007/S10489-022-03867-9},
    issn = {15737497}
}

@article{Gao2024DeepFakeContent,
    title = {{DeepFake detection based on high-frequency enhancement network for highly compressed content}},
    year = {2024},
    journal = {Expert Systems with Applications},
    author = {Gao, Jie and Xia, Zhaoqiang and Marcialis, Gian Luca and Dang, Chen and Dai, Jing and Feng, Xiaoyi},
    month = {9},
    pages = {123732},
    volume = {249},
    publisher = {Pergamon},
    url = {https://www.sciencedirect.com/science/article/pii/S0957417424005980?via%3Dihub},
    doi = {10.1016/J.ESWA.2024.123732},
    issn = {0957-4174}
}

@article{Yan2023DeepfakeBench:Detection,
    title = {{DeepfakeBench: A Comprehensive Benchmark of Deepfake Detection}},
    year = {2023},
    journal = {Advances in Neural Information Processing Systems},
    author = {Yan, Zhiyuan and Zhang, Yong and Yuan, Xinhang and Lyu, Siwei and Wu, Baoyuan},
    month = {7},
    volume = {36},
    publisher = {Neural information processing systems foundation},
    url = {https://arxiv.org/pdf/2307.01426},
    isbn = {9781713899921},
    issn = {10495258},
    arxivId = {2307.01426}
}

@article{Lin2024ExploitingDetection,
    title = {{Exploiting Facial Relationships and Feature Aggregation for Multi-Face Forgery Detection}},
    year = {2024},
    journal = {IEEE Transactions on Information Forensics and Security},
    author = {Lin, Chenhao and Yi, Fangbin and Wang, Hang and Deng, Jingyi and Zhao, Zhengyu and Li, Qian and Shen, Chao},
    pages = {8832--8844},
    volume = {19},
    publisher = {Institute of Electrical and Electronics Engineers Inc.},
    doi = {10.1109/TIFS.2024.3461469},
    issn = {15566021},
    arxivId = {2310.04845}
}

@article{Alanazi2024ExploringCountermeasures,
    title = {{Exploring deepfake technology: creation, consequences and countermeasures}},
    year = {2024},
    journal = {Human-Intelligent Systems Integration 2024 6:1},
    author = {Alanazi, Sami and Asif, Seemal},
    number = {1},
    month = {9},
    pages = {49--60},
    volume = {6},
    publisher = {Springer},
    url = {https://link.springer.com/article/10.1007/s42454-024-00054-8},
    isbn = {0123456789},
    doi = {10.1007/S42454-024-00054-8},
    issn = {2524-4884}
}

@article{Nguyen2024ExploringAnalysis,
    title = {{Exploring Self-Supervised Vision Transformers for Deepfake Detection: A Comparative Analysis}},
    year = {2024},
    journal = {ArXiV},
    author = {Nguyen, Huy H. and Yamagishi, Junichi and Echizen, Isao},
    month = {8},
    url = {https://arxiv.org/pdf/2405.00355v2},
    arxivId = {2405.00355v2}
}

@article{Rossler2019FaceForensics++:Images,
    title = {{FaceForensics++: Learning to Detect Manipulated Facial Images}},
    year = {2019},
    journal = {Proceedings of the IEEE International Conference on Computer Vision},
    author = {Rossler, Andreas and Cozzolino, Davide and Verdoliva, Luisa and Riess, Christian and Thies, Justus and Niessner, Matthias},
    month = {1},
    pages = {1--11},
    publisher = {Institute of Electrical and Electronics Engineers Inc.},
    url = {https://arxiv.org/pdf/1901.08971},
    isbn = {9781728148038},
    doi = {10.1109/ICCV.2019.00009},
    issn = {15505499},
    arxivId = {1901.08971}
}

@article{Tanfoni2024FacialApproach,
    title = {{Facial Segmentation in Deepfake Classification: a Transfer Learning Approach}},
    year = {2024},
    journal = {Procedia Computer Science},
    author = {Tanfoni, Marco and Ceroni, Elia Giuseppe and Pancino, Niccolò and Bianchini, Monica and Maggini, Marco},
    number = {C},
    month = {1},
    pages = {4160--4168},
    volume = {246},
    publisher = {Elsevier},
    url = {https://www.sciencedirect.com/science/article/pii/S1877050924022749?via%3Dihub},
    doi = {10.1016/J.PROCS.2024.09.255},
    issn = {1877-0509}
}

@article{Cai2023GlitchLocalization,
    title = {{Glitch in the matrix: A large scale benchmark for content driven audio–visual forgery detection and localization}},
    year = {2023},
    journal = {Computer Vision and Image Understanding},
    author = {Cai, Zhixi and Ghosh, Shreya and Dhall, Abhinav and Gedeon, Tom and Stefanov, Kalin and Hayat, Munawar},
    month = {11},
    pages = {103818},
    volume = {236},
    publisher = {Academic Press},
    url = {https://www.sciencedirect.com/science/article/pii/S1077314223001984},
    doi = {10.1016/J.CVIU.2023.103818},
    issn = {1077-3142},
    arxivId = {2305.01979}
}

@article{Le2021OpenForensics:V.1.0.0,
    title = {{OpenForensics: Multi-Face Forgery Detection And Segmentation In-The-Wild Dataset [V.1.0.0]}},
    year = {2021},
    journal = {Zenodo},
    author = {Le, Trung-Nghia and Nguyen, Huy H. and Yamagishi, Junichi and Echizen, Isao},
    month = {10},
    url = {https://zenodo.org/records/5528418},
    doi = {10.5281/ZENODO.5528418},
    language = {English}
}

@article{Yu2019PDA:Networks,
    title = {{PDA: Progressive Data Augmentation for General Robustness of Deep Neural Networks}},
    year = {2019},
    journal = {ArXiV},
    author = {Yu, Hang and Liu, Aishan and Liu, Xianglong and Li, Gengchao and Luo, Ping and Cheng, Ran and Yang, Jichen and Zhang, Chongzhi},
    month = {9},
    url = {https://arxiv.org/pdf/1909.04839},
    arxivId = {1909.04839}
}

@article{Dolhansky2020TheDataset,
    title = {{The DeepFake Detection Challenge (DFDC) Dataset}},
    year = {2020},
    journal = {ArXiV},
    author = {Dolhansky, Brian and Bitton, Joanna and Pflaum, Ben and Lu, Jikuo and Howes, Russ and Wang, Menglin and Ferrer, Cristian Canton},
    month = {6},
    url = {https://arxiv.org/pdf/2006.07397},
    isbn = {2006.07397v4},
    arxivId = {2006.07397}
}

@article{Guarnera2022TheChallenge,
    title = {{The Face Deepfake Detection Challenge}},
    year = {2022},
    journal = {Journal of Imaging},
    author = {Guarnera, Luca and Giudice, Oliver and Guarnera, Francesco and Ortis, Alessandro and Puglisi, Giovanni and Paratore, Antonino and Bui, Linh M.Q. and Fontani, Marco and Coccomini, Davide Alessandro and Caldelli, Roberto and Falchi, Fabrizio and Gennaro, Claudio and Messina, Nicola and Amato, Giuseppe and Perelli, Gianpaolo and Concas, Sara and Cuccu, Carlo and Orr{\`{u}}, Giulia and Marcialis, Gian Luca and Battiato, Sebastiano},
    number = {10},
    month = {10},
    pages = {263},
    volume = {8},
    publisher = {MDPI},
    url = {https://pmc.ncbi.nlm.nih.gov/articles/PMC9605671/},
    doi = {10.3390/JIMAGING8100263},
    issn = {2313433X},
    pmid = {36286357}
}

@article{Ojha2023TowardsModels,
    title = {{Towards Universal Fake Image Detectors that Generalize Across Generative Models}},
    year = {2023},
    journal = {Proceedings of the IEEE Computer Society Conference on Computer Vision and Pattern Recognition},
    author = {Ojha, Utkarsh and Li, Yuheng and Lee, Yong Jae},
    month = {2},
    pages = {24480--24489},
    volume = {2023-June},
    publisher = {IEEE Computer Society},
    url = {https://arxiv.org/pdf/2302.10174},
    isbn = {9798350301298},
    doi = {10.1109/CVPR52729.2023.02345},
    issn = {10636919},
    arxivId = {2302.10174}
}

@article{Touvron2020TrainingAttention,
    title = {{Training data-efficient image transformers {\&} distillation through attention}},
    year = {2020},
    journal = {Proceedings of Machine Learning Research},
    author = {Touvron, Hugo and Cord, Matthieu and Douze, Matthijs and Massa, Francisco and Sablayrolles, Alexandre and J{\'{e}}gou, Hervé},
    month = {12},
    pages = {10347--10357},
    volume = {139},
    publisher = {ML Research Press},
    url = {https://arxiv.org/pdf/2012.12877},
    isbn = {9781713845065},
    issn = {26403498},
    arxivId = {2012.12877}
}

@article{Yan2024TranscendingDetection,
    title = {{Transcending Forgery Specificity with Latent Space Augmentation for Generalizable Deepfake Detection}},
    year = {2024},
    journal = {ArXiV},
    author = {Yan, Zhiyuan and Luo, Yuhao and Lyu, Siwei and Liu, Qingshan and Wu, Baoyuan},
    month = {3},
    url = {https://arxiv.org/pdf/2311.11278v2},
    arxivId = {2311.11278v2}
}

@article{Cocchi2023UnveilingAnalysis,
    title = {{Unveiling the Impact of Image Transformations on Deepfake Detection: An Experimental Analysis}},
    year = {2023},
    journal = {Lecture Notes in Computer Science (including subseries Lecture Notes in Artificial Intelligence and Lecture Notes in Bioinformatics)},
    author = {Cocchi, Federico and Poppi, Samuele and Cornia, Marcella and Baraldi, Lorenzo and Cucchiara, Rita},
    month = {1},
    pages = {345--356},
    volume = {14234 LNCS},
    publisher = {Springer, Cham},
    url = {https://link.springer.com/chapter/10.1007/978-3-031-43153-1_29},
    isbn = {978-3-031-43153-1},
    doi = {10.1007/978-3-031-43153-1{\_}29},
    issn = {1611-3349}
}

@article{Zi2021WildDeepfake:Detection,
    title = {{WildDeepfake: A Challenging Real-World Dataset for Deepfake Detection}},
    year = {2021},
    journal = {MM 2020 - Proceedings of the 28th ACM International Conference on Multimedia},
    author = {Zi, Bojia and Chang, Minghao and Chen, Jingjing and Ma, Xingjun and Jiang, Yu Gang},
    month = {1},
    pages = {2382--2390},
    publisher = {Association for Computing Machinery, Inc},
    url = {https://arxiv.org/pdf/2101.01456},
    isbn = {9781450379885},
    doi = {10.1145/3394171.3413769},
    arxivId = {2101.01456}
}




% %% For authoryear reference style

% \end{thebibliography}
\end{document}